\documentclass[sigconf,natbib=true]{acmart}

\AtBeginDocument{%
  \providecommand\BibTeX{{%
    \normalfont B\kern-0.5em{\scshape i\kern-0.25em b}\kern-0.8em\TeX}}}

\setcopyright{acmcopyright}
\copyrightyear{2023}
\acmYear{2023}
\acmDOI{XXXXXXX.XXXXXXX}
\acmConference[SIGIR'23]{Proceedings of The Web Conference
2023 (SIGIR'23)}{July 23--27, 2023}{Taipei}
\acmBooktitle{Proceedings of the SIGIR'23,
  July 23--27, 2023, Taipei}
\acmPrice{15.00}
\acmISBN{978-1-4503-XXXX-X/18/06}

\usepackage{amssymb}
\floatname{algorithm}{Algorithm}
\usepackage[ruled,linesnumbered]{algorithm2e}
\usepackage[ruled]{algorithm2e}
\usepackage{algorithmic}
\usepackage{subcaption}
\usepackage{graphicx}

\usepackage{diagbox}
\usepackage{url}
\usepackage{booktabs}
\usepackage{threeparttable}
 \usepackage{multirow}

\begin{document}

\title{DREAM: Adaptive Reinforcement Learning based on Attention Mechanism for Temporal Knowledge Graph Reasoning}

\author{Shangfei Zheng}
\affiliation{%
  \institution{School of Computer Science and Technology, Soochow University}
  \city{Suzhou}
  \country{China}
}
\email{sfzhengsuda@stu.suda.edu.cn}

\author{Hongzhi Yin}
\authornotemark[1]
\affiliation{%
  \institution{School of Information Technology and Electrical Engineering, \\The University of Queensland}
  \city{Brisbane}
  \country{Australia}
}
\email{db.hongzhi@gmail.com}

\author{Tong Chen}
\affiliation{%
  \institution{School of Information Technology and Electrical Engineering,\\ The University of Queensland}
  \city{Brisbane}
  \country{Australia}
}
\email{tong.chen@uq.edu.au}

\author{Quoc Viet Hung Nguyen}
\affiliation{%
  \institution{School of Information and Communication Technology, \\ Griffith University}
  \city{Gold Coast}
  \country{Australia}
}
\email{henry.nguyen@griffith.edu.au}

\author{Wei Chen}

\affiliation{%
  \institution{School of Computer Science and Technology, Soochow University}
  \city{Suzhou}
  \country{China}
}
\email{robertchen@suda.edu.cn}

\author{Lei Zhao}
\authornote{Corresponding author}
\affiliation{%
  \institution{School of Computer Science and Technology, Soochow University}
  \city{Suzhou}
  \country{China}
}
\email{zhaol@suda.edu.cn}



\begin{abstract}
  Temporal knowledge graphs (TKGs) model the temporal evolution of events and have recently attracted increasing attention. Since TKGs are intrinsically incomplete, it is necessary to reason out missing elements. Although existing TKG reasoning methods have the ability to predict missing future events, they fail to generate explicit reasoning paths and lack explainability. As reinforcement learning (RL) for multi-hop reasoning on traditional knowledge graphs starts showing superior explainability and performance in recent advances, it has opened up opportunities for exploring RL techniques on TKG reasoning.  However, the performance of RL-based TKG reasoning methods is limited due to: (1) lack of ability to capture temporal evolution and semantic dependence  jointly;  (2) excessive reliance on manually designed rewards. To overcome these challenges, we propose an a\underline{d}aptive \underline{r}einforcement l\underline{e}arning model based on \underline{a}ttention \underline{m}echanism (DREAM) to predict missing elements in the future. Specifically, the model contains two components: (1) a multi-faceted attention representation learning method that captures  semantic dependence and temporal evolution jointly; (2) an adaptive RL framework that conducts multi-hop reasoning by adaptively learning the reward functions. 
  Experimental results demonstrate DREAM outperforms state-of-the-art models on public datasets. 

\end{abstract}






\maketitle

\section{Introduction}
A knowledge graph (KG) stores a large variety of information about the real world, and has achieved great success in many downstream applications, such as question answering \cite{RS5} and  recommendation systems \cite{RS3, RS6, RS4}. Traditional KG is expressed in terms of static relation triplets (\emph{$e_s$}, \emph{r}, \emph{$e_d$}), where \emph{$e_s$} and $e_d$ are respectively subject and object entities, and \emph{r} represents a relation, e.g., (\emph{COVID-19}, \emph{Occur}, \emph{City Hall}). However, events constantly exhibit complex temporal dynamics in the real world \cite{kgsurveytkde, RS2}. To reflect the temporal nature of events, traditional KGs are extended to the temporal knowledge graphs (TKGs) that additionally associate relation triplets with a timestamp (\emph{$e_s$}, \emph{r}, \emph{$e_d$}, \emph{t}), e.g., (\emph{COVID-19}, \emph{Occur}, \emph{City Hall}, 2022--12--6).

\begin{figure*}
 \centering 
  \label{Figure 1}
  \includegraphics[width=0.9\linewidth,height=5.3cm]{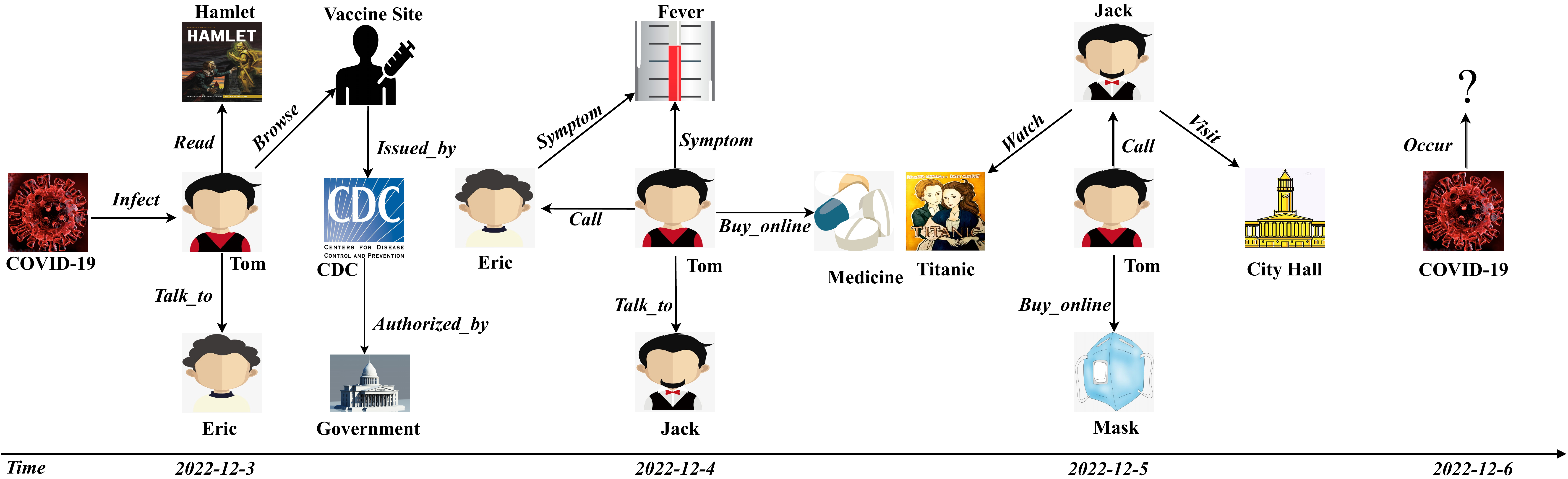}
 \hfill 
 \vspace{-0.25cm}
 \caption{
A small fragment of a TKG. A sequence of snapshots containing entities and relations is arranged by ascending timestamp. By utilizing historical and semantic information, (\emph{COVID-19}, \emph{Occur}, \emph{City Hall}, 2022--12--6) can be inferred.
}
\vspace{-0.35cm}
\end{figure*}

Given the intrinsic incompleteness of TKGs, recent temporal knowledge graph reasoning (TKGR) studies have contributed to inferring missing events from known ones \cite{kgr_survey}. For example, the missing element \emph{City Hall} in the quadruple query (\emph{COVID-19}, \emph{Occur}, ?, 2022--10--6) can be predicted via TKGR. Given a TKG with a time span between \emph{$t_0$} and \emph{$t_T$}, TKGR methods can be categorized into two types, i.e., interpolation and extrapolation \cite{RE-GCN, RE-NET}. The former aims to reason out missing facts for observed timestamp \emph{t} that is not greater than \emph{$t_T$} (i.e., \emph{$t_0$} $\leq$ \emph{t} $\leq$  \emph{$t_T$}). In contrast, the latter focuses on forecasting new facts over future timestamps \emph{t} $\textgreater$ \emph{$t_T$}. Extrapolation has high values in practical applications and is helpful in many fields, such as financial risk control \cite{financialapp}, epidemic prediction \cite{Cluster}, and disaster warning  \cite{disaster}. In addition, predictions of future events have become an essential topic in AI research \cite{AIwork, AI-research}, making extrapolation-based approaches the mainstream of current TKGR research \cite{tkgc-survey}  as well as the focal point of our work.

In the literature, there has been a long line of studies in extrapolated TKGR, such as tensor decomposition-based methods \cite{TNTComplEx, tensorc}, translation-based methods \cite{TTransE, TA-transE}, neural network-based methods \cite{RE-GCN, RE-NET, CyGNet}, etc. While these studies have leveraged temporal information to predict missing elements over TKGs, they act as black-box systems that lack explainability \cite{THML, RLH}. Recent advances in traditional knowledge graph reasoning have shown that reinforcement learning (RL)-based multi-hop reasoning achieves promising  and interpretable results \cite{kgsurvey, RS1}. This success has inspired using RL on extrapolated TKGR. Representatively, TPath \cite{TPath}, TAgent\cite{TAgent}, TITer \cite{TiTer}, and CluSTeR \cite{Cluster} learn multi-hop explainable paths by leveraging the symbolic combination and transmission of relations. For example, by connecting (\emph{COVID-19}, {\emph{Infect}}, \emph{Tom}, 2022--12--3),  (\emph{Tom},  \emph{Talk\_to},  \emph{Jack}, 2022--12--4) and (\emph{Jack}, \emph{Visit}, \emph{City Hall}, 2022--12--5), RL-based models can reason out a quadruple (\emph{COVID-19}, \emph{Occur}, \emph{City Hall}, 2022--12--6).  Although the above methods can conduct multi-hop reasoning in an explainable manner over TKGs, their reasoning performance is restricted due to the following two \emph{challenges}.


\textbf{Challenge \uppercase\expandafter{\romannumeral1}}. How to capture both temporal evolution and semantic  dependence simultaneously has not yet been explored. 
Some RL-based TKGR methods \cite{TPath, TAgent,TiTer} are powerless to capture the \textbf{temporal evolution} (i.e., the dynamic influence of historical events). Accordingly,  CluSTeR \cite{Cluster} considers historical events in the temporal reasoning stage, but only utilizes limited historical information from a small look-back window (events within the last three days) by employing recurrent neural networks (RNNs). This is because the recurrent method scales poorly with an increasing number of timestamps \cite{attention}. In addition, RNN-based methods commonly require a large amount of training data which limits the performance of CluSTeR \cite{RNNsurvey}. Furthermore, the \textbf{semantic dependence} is largely overlooked by existing RL-based TKGR methods. In fact, an entity plays diverse roles in different relations \cite{ATTNPath}. Take Figure 1 as an example, $Tom$ may be connected with hobby relations like $Read$ and also clinical relations like \emph{Infected}, which are not equally informative when responding to query (\emph{COVID-19}, \emph{Occur}, ?, 2022--10--6). Paying more attention to relations highly correlated with query relations is beneficial for reducing semantic noise in the reasoning process \cite{renvid}. Besides, the semantic relevance between an entity and its neighbors tends to be negatively correlated with their distance \cite{GMH}. 
For instance, the possibility that Eric's 4-hop neighbor \emph{Government} provides reasoning clues is lower than their  2-hop neighbor \emph{COVID-19}. Hence, it is arguably necessary to exploit both temporal evolution and semantic dependence simultaneously to improve reasoning performance.

\textbf{Challenge \uppercase\expandafter{\romannumeral2}}. The second challenge is that the aforementioned RL-based  methods heavily rely on manually designed rewards, which easily lead to \emph{sparse reward dilemmas}, \emph{laborious design process} and \emph{performance fluctuation} \cite{rewardRL}. Any of the three limits directly impedes the performance and generalizability of RL-based TKGR models. 
(1) \emph{Sparse reward dilemmas}. The sparse reward resulting from considering only terminal reward leads to slow or even failed learning \cite{MultiHop, rewardRLcon}. For example, TAgent is subject to convergence failure since its reward is a binary terminal reward \cite{TAgent}.
(2) \emph{Laborious design process}. Experienced experts are constantly engaged to design some candidate auxiliary rewards that will potentially improve the reasoning performance, and then carefully select an optimal reward function (e.g., CluSTeR adds an auxiliary beam-level reward  \cite{Cluster}). Unfortunately, this meticulous artificial design with domain expertise has low generalizability across datasets \cite{generaRL, RLregener}. (3) \emph{Performance fluctuation}. Manually designed functions are difficult to balance exploration and exploitation, which incurs inevitable fluctuation of reasoning performance and decision bias of RL \cite{RLbalan, RLrewardBal}. Referring to the experiment of the model TPath \cite{TPath}, all evaluation indicators fluctuate significantly on different TKG datasets, where a major cause is the decision bias triggered by the manually designed path diversity reward.

In light of the aforementioned challenges in TKGR, we propose a novel model entitled \textbf{DREAM} (a\underline{d}aptive \underline{r}einforcement l\underline{e}arning model based on \underline{a}ttention \underline{m}echanism). Our innovations are  inspired by two points: (1) attention mechanism that is more capable of incorporating diverse historical information into semantic representations than recurrent methods \cite{attention}; (2) generative  adversarial imitation learning that adaptively learns policy from expert demonstrations \cite{GAIL}. Notably, the main difference between DREAM and existing multi-hop reasoning models is that our model not only elegantly captures both temporal evolution and semantic dependence but also conducts multi-hop reasoning via  a novel adaptive reinforcement learning framework with an adaptive reward function. Specifically, the model contains two components. (1) To solve \textbf{Challenge \uppercase\expandafter{\romannumeral1}}, a multi-faceted attention representation (MFAR) method is proposed to preserve both the semantic and temporal properties of TKGs. Its relation-aware attenuated graph attention module explores the semantic dependence of TKGs from multi-hop entities and relations. A temporal self-attention module captures evolution over multiple time steps by flexibly weighting historical context.  (2) To solve \textbf{Challenge \uppercase\expandafter{\romannumeral2}}, we design an adaptive reinforcement learning  framework (ARLF) based on generative adversarial imitation learning. The goal of ARLF is to predict the missing elements by learning adaptive rewards at both semantic and temporal rule levels. To sum up, the contributions of this paper are as follows.

\begin{itemize}
    \item To the best of our knowledge, we are the first to point out the necessity of jointly exploiting temporal evolution and semantic dependence, as well as the need for designing adaptive reward functions when predicting missing elements in RL-based TKGR methods.

   \item We propose DREAM, an effective and  explainable model to conduct multi-hop reasoning by leveraging the symbolic compositionality of relations. In DREAM, we propose MFAR to learn latent feature representations of TKGs. These features are fed into ARLF, a novel RL-based framework with minimal dependence on handcrafted reward functions, to achieve
  explainable reasoning results.

   \item We conduct extensive experiments on four benchmark datasets. The results showcase the superiority of DREAM in comparison with state-of-the-art baselines.
\end{itemize}


\section{Related Work}
\subsection{Traditional Knowledge Graph  Reasoning} 
A KG is a multi-relational graph that reflects the internal connections of events in reality \cite{kgsurvey}. Traditional KGs are represented by triplets consisting of entities and relations. Whether manually constructed or automatically extracted, KGs are incomplete \cite{MultiHop, tkgc-survey}.  
There has been a wide range of KG reasoning approaches to address the shortage of incompleteness  \cite{kgr_survey}, such as translation-based method \cite{TransE}, neural network-based method \cite{kgrnn} and tensor decomposition-based method \cite{decompkgr}, etc. However, the above methods are only suitable for single-hop reasoning by modeling one-step relations and lack explainability \cite{RLH}. State-of-the-art KG reasoning methods are known to be primarily multi-hop reasoning based on RL \cite{MINERVA}. These methods explore the semantic composition of multi-hop relations, which naturally forms the  explainable inferred results \cite{kgr_survey}. Typical RL-based multi-hop reasoning methods include THML \cite{THML}, DeepPath \cite{DeepPath}, MultiHop \cite{MultiHop}, etc.  
Additionally, rule-based reasoning methods \cite{NeuralLP, RNNLogic} also perform explainable reasoning by learning the logic rules. Despite the great efforts made by these studies, they are not  designed for TKGR \cite{RE-NET, TiTer}.

\subsection{Temporal Knowledge Graph Reasoning} 

TKGR methods are usually categorized into interpolation and extrapolation \cite{tkgc-survey}. The former infers missing elements at historical timestamps \cite{TTransE, TA-transE, TNTComplEx, DE-SimplE}.  The latter setting is orthogonal to our work, which attempts to infer missing facts in the future. Typically, the performance of interpolated TKGR models is lower than that of extrapolated ones when predicting future events  \cite{TTransE, TNTComplEx, TA-transE}.
For extrapolation,  some studies \cite{Know-Evolve, Hawkes, DyRep, TANGO} adopt the temporal point process or continuous-time dynamic embeddings to estimate the conditional probability in continuous time. However, these methods cannot make full use of the structural information \cite{RE-GCN}. By integrating convolutional networks and curriculum learning, CEN addresses the challenge of structure-variability evolutional patterns \cite{CEN}. In addition, RE-NET \cite{RE-NET} and RE-GCN \cite{RE-GCN} utilize RNNs to learn the historical property under different timestamps. Similarly,  TiRGN \cite{TiRGN} and HiSMatch \cite{HiSMatch} respectively employ RNNs and structure encoders to learn historical patterns. Nonetheless, these TKGR methods not only fail to jointly capture temporal evolution and semantic dependence but also scale poorly due to the limitation of RNNs \cite{attengraph}. 


Another class of TKGR methods based on RL is most relevant to our work, which outputs explainable predicted results under the lack of multi-faceted representation. Formally, RL-based TKGR methods treat TKGR as a Markov decision process, which aims to learn optimal policy about inferring missing elements in TKGs. Specifically, TAgent \cite{TAgent} only adopts binary terminal rewards to conduct TKGR so that it cannot obtain sufficient rewards \cite{RLrewardBal}. To further improve the quality of the reward function, TPath \cite{TPath} and TITer \cite{TiTer} add path diversity reward and time-shaped reward, respectively. CluSTeR employs an RNN to learn structural information and integrates the information into the beam-level reward function \cite{Cluster}. However, the performance of these models relies on manually designed rewards, hence is limited due to sparse reward dilemmas, laborious design process, and performance fluctuation. Although graph learning-based xERTE \cite{XERTE} and rule-based TLogic \cite{TLogic} also conduct explainable reasoning, their reasoning performance is generally lower than that of SOTA reasoning  models based on RL.

\begin{figure*}
 \centering 
  \label{Figure 2}
  \includegraphics[width=0.9\linewidth,height=6.8cm]{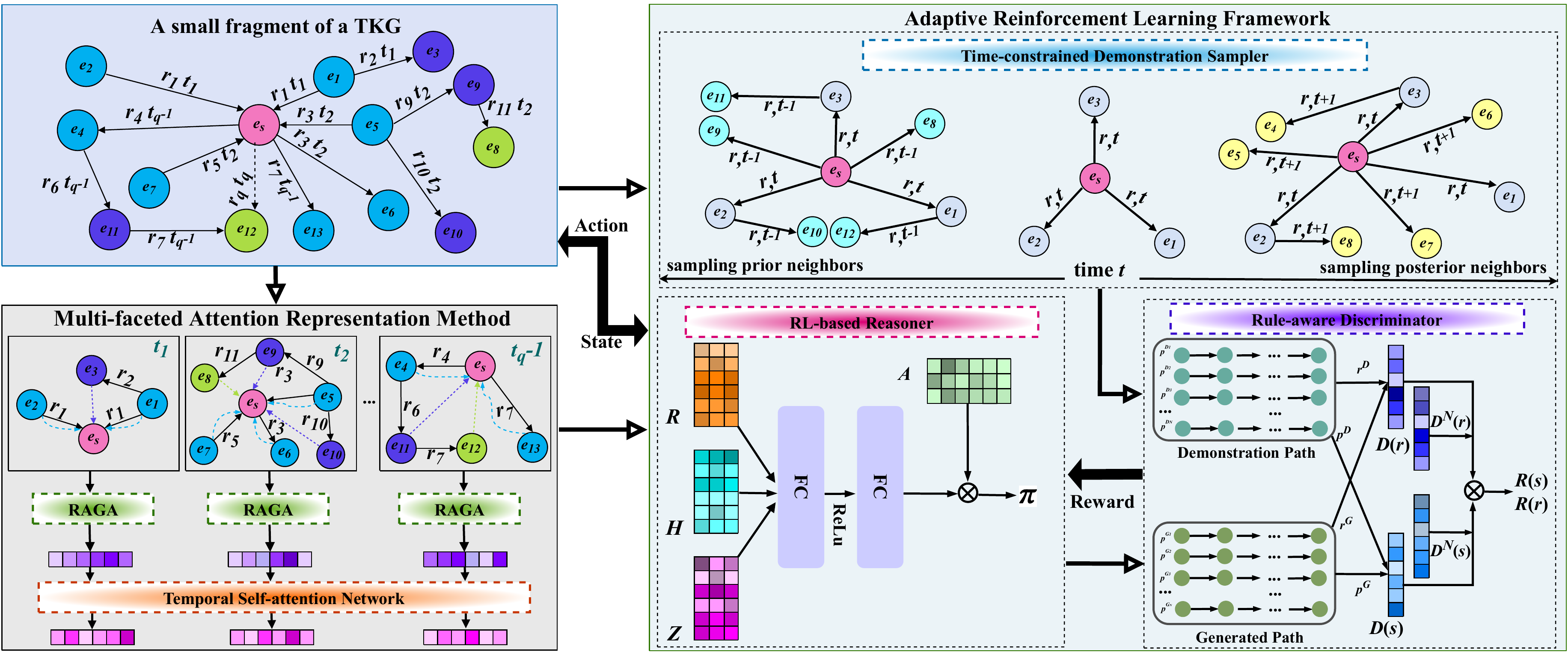}
  \vspace{-3.5pt}
 \hfill 
 \caption{The detailed schematic diagram of DREAM. 
An RL-based reasoner outputs reasoning paths after representations are fed to it. Next, a sampler obtains  demonstration paths by bi-directional sampling. Then, the discriminator generates adaptive rewards for the reasoner. Finally, this reasoner updates reasoning policies and interacts with TKG to complete the prediction.}
\vspace{-7pt}
\end{figure*}

\subsection{Reinforcement Learning}
Most RL-based algorithms utilize an agent to get rewards from the environment and optimize policies by Markov decision process (MDP) \cite{RLH, MMKGR}. A key challenge is to construct reward functions when applying RL to real-world problems \cite{rewardRL}. Actually, manually designed rewards tend to suffer from sparse reward dilemmas in RL-based models, which causes slow or even failed learning \cite{MultiHop, rewardRLcon}. To alleviate the above problems, reward shaping has received extensive attention \cite{generaRL}. However, the design principles of these laborious methods still rely on human experience excessively, which results in low generalizability of different datasets and performance fluctuations \cite{RLregener, RLrewardBal}. To solve the above problems, inverse reinforcement learning (IRL) tries to learn the rewards from expert examples adaptively, but the learning loop of IRL results in a large running cost in large environments \cite{IRL}. Consequently, generative adversarial imitation learning is designed in game simulation fields to eliminate any intermediate IRL steps and imitate expert demonstrations via the generative adversarial network (GAN). 


\section{Preliminaries}

A TKG $\mathcal{G}$ = ($\mathcal{E}$, $\mathcal{R}$, $\mathcal{T}$, $\mathcal{Q}$) is a directed multi-relation  graph with time-stamped edges between entities, where $\mathcal{E}$, $\mathcal{R}$ and $\mathcal{T}$ are the set of entities, relations and timestamps, respectively. $\mathcal{Q}$ =  \{ (\emph{$e_s$},
\emph{$r_q$}, \emph{$e_d$}, $t_q$) $\mid$ \emph{$e_s$}, \emph{$e_d$} $\in$ $\mathcal{E}$, \emph{$r_q$} $\in$ $\mathcal{R}$, \emph{$t_q$} $\in$  $\mathcal{T}$\} is a set of quadruples in $\mathcal{G}$, where $e_s$, $e_d$, $r_q$ and $t_q$ are a source entity, a target  entity, the relation between them, and a timestamp, respectively. A TKG is regarded as \emph{a sequence of snapshots} ordered ascendingly based on their timestamps, i.e., $\mathcal{G}$ = \{{$\mathcal{G}_1$, $\mathcal{G}_2$,...,$\mathcal{G}_T$}\}. Following \cite{Cluster, TiTer}, inverse relation is also appended to the quadruple, i.e., each quadruple (\emph{$e_s$}, \emph{$r$}, \emph{$e_d$}, $t$) is equivalent to the quadruple (\emph{$e_d$}, \emph{$r^{-1}$}, \emph{$e_s$}, $t$). Without loss of generality, we infer the missing subject entity by converting (?, \emph{$r_q$}, \emph{$e_d$}, $t_q$) to (\emph{$e_d$}, \emph{$r^{-1}_{q}$}, ?, $t_q$). The problem studied in this paper is extrapolated TKGR, which is formalized as link prediction that tries to infer the future quadruple. Formally, for a quadruple query (\emph{$e_s$}, \emph{$r_q$},$?$,\emph{$t_q$}), the goal of extrapolated TKGR is to predict the missing entity \emph{$e_d$} given the set of previous  facts before \emph{$t_q$}, denoted as \{ {$\mathcal{G}_1$, $\mathcal{G}_2$,...,$\mathcal{G}_{t_{q}-1}$} \}, via generating an explainable \emph{k}-hop reasoning path.
For instance, the object entity of the quadruple query (\emph{COVID-19}, \emph{Occur}, ?, 2022--12--6) is missing. By connecting the 3-hop path \emph{COVID-19} $\xrightarrow[2022-10-3]{Infect}$  \emph{Tom} $\xrightarrow[2022-10-4]{Talk\_{to}}$ \emph{Jack} $\xrightarrow[2022-10-5]{Visit}$ \emph{City Hall}, the RL-based model obtains  a quadruple (\emph{COVID-19}, \emph{Occur}, \emph{City Hall}, 2022--12--6). We formulate the TKG problem  as follows:
\begin{itemize}
\item \emph{Input is a quadruple query (\emph{$e_s$}, \emph{$r_q$},$?$,\emph{$t_q$}) or (?, \emph{$r_q$},\emph{$e_d$},\emph{$t_q$}), where $r_q$ is a query  relation, "?" is the missing target or source entity, \emph{$t_q$} is a future timestamp (i.e., it only appears in the test set)}.
\item \emph{Output is the inferred entity acquired via a reasoning path with no more than k hops, k > 1.}
\end{itemize}

\section{Methodology}
\subsection{Overview of DREAM}

As shown in Figure 2, DREAM contains two components: (1) a multi-faceted attention representation (MFAR), which captures both temporal evolution and semantic dependence simultaneously and generates multi-faceted representation; (2) an adaptive reinforcement learning framework (ARLF) with an adaptive  reward mechanism, which conducts explainable reasoning based on the representation obtained from MFAR. In what follows, we unfold the design of both components.

\subsection{Multi-faceted Attention Representation}
Existing methods are powerless to capture both temporal evolution and semantic dependence simultaneously, which limits the utilization of temporal data \cite{bothsurvey, TKG-survey-bai}. To address the problem, we introduce a multi-faceted attention representation method. MFAR jointly learns entity and relation features in multi-hop neighbors, and captures potential temporal evolution by flexibly weighting historical events. Notably, the technical differences from the existing representation learning methods in the field of TKGR lie in the following two points. (1) We extend Graph Attention Network (GAT) by calculating entities and relations within multi-hop neighborhoods, while GAT only considers entity features in a one-hop neighborhood. Moreover, compared with embedding methods based on relation-aware graph networks in traditional 
 KGs \cite{RGHAT, M2GNN}, our method not only introduces an attenuated mechanism to simulate the declining contribution of multi-hop entities \cite{THML}, but also has lower computational cost. This is because MFAR directly associates multi-hop neighbors by summing the embeddings of relational paths, rather than increasing the number of network layers or introducing algebraic space  \cite{cost}. (2) We capture reasoning clues from a longer range of previous timestamps  by learning attentive weights rather than using a non-scalable recurrent method \cite{attention}. 


Specifically, MFAR decouples graph attention and temporal attention into independent modules, i.e., \emph{relation-aware attenuated graph attention} (RAGA) and \emph{temporal self-attention network} (TSAN). After obtaining static representations of all entities and relations, RAGA extracts multi-hop neighborhood information through an attenuated mechanism in each timestamp. Then, TSAN captures long-range temporal evolution and depends on the historical information of each entity. This decoupled design pattern facilitates efficient  and flexible parallelism in  sequential tasks \cite{decouple}. 

\subsubsection{Relation-aware Attenuated Graph Attention}

Typically, existing TKGR methods \cite{TiTer, Cluster, RE-GCN, TiRGN, CEN, HiSMatch} learn structural information by solely focusing on representations of entities and relations in a disjoint manner, which cannot exploit the diverse semantics of event triplets \cite{TKG-survey-bai}. To make matters worse, existing methods introduce noise into the reasoning process due to the neglect of semantic decay \cite{MultiHop}. Inspired by \cite{NathaniCSK19}, we propose RAGA to solve the above problems of semantic dependencies in this subsection.  The input of RAGA is  $\mathcal{G}_t$ ($t$<$t_q$) containing a set of entity representations \{$\textbf{\emph{e}}_i$ $\in$ $\mathbb{R}^{D}$, ${\forall}$$e_i$ $\in$ $\mathcal{E}$\} and relation representations \{$\textbf{\emph{r}}_i$ $\in$ $\mathbb{R}^{F'}$, ${\forall}$$r_i$ $\in$ $\mathcal{R}$\}. The output is a set of new entity representations   \{$\textbf{\emph{e}}_i^{s}$ $\in$ $\mathbb{R}^{D_e^{'}}$, ${\forall}$$e_i$ $\in$ $\mathcal{E}$\} with $D_e^{'}$ dimensions, which fully captures semantic dependencies in  $\mathcal{G}_t$. 
 

Specifically, to obtain the direct embedding from a given entity $e_i$ to its $k$-hop neighbor entity, we randomly sample a shortest path between $e_i$ and its $k$-hop neighbor entity \cite{DeepPath} and then construct an auxiliary relation embedding $\textbf{\emph{r}}_{m}$ by performing a summation of embeddings of all relations in this path, i.e., $\textbf{\emph{r}}_{m}$ = $\textbf{\emph{r}}_{1}$+...+$\textbf{\emph{r}}_{k}$. The given relation $r_k$ is denoted as a relation of $e_i$ to a neighbor entity in the timestamp $\mathcal{G}_t$, and represented as follows, 
\begin{equation}
 \textbf{\emph{r}}_{k}=
\begin{cases}
\quad \textbf{\emph{r}} & \text{\emph{k} = 1}\\
\quad \textbf{\emph{r}}_{m}&  \text{\emph{k} > 1}
\end{cases}
\end{equation}
where \emph{k} denotes the number of hops in the relation path, $\textbf{\emph{r}}$ is the relation embedding between $e_i$ and direct  neighbor entities. 

We first obtain the vector representation of each triple associated with entity $e_i$ by performing concatenation, which can model the difference of roles played by an entity in different relations in a given snapshot $\mathcal{G}_t$. The calculation process is as follows, 
\begin{equation}
\textbf{\emph{t}}_{ikj} = \textbf{W}_1[\textbf{\emph{e}}_{i}\oplus\textbf{\emph{r}}_{k}\oplus\textbf{\emph{e}}_{j}]
\end{equation}
where $\oplus$ is the concatenation, $\textbf{\emph{t}}_{ikj}$ is the triple vector representation in a given snapshot $\mathcal{G}_t$, $\textbf{\emph{e}}_{j}$ represents the embedding of a direct in-flowing neighbor or multi-hop neighbor of the entity $e_i$ in $\mathcal{G}_t$. 

Next, to model the distance-sensitive contribution of $k$-hop neighbor entities of a given entity $e_i$ in $\mathcal{G}_t$, we define a Gaussian-based attenuated coefficient due to the stability and simplicity of the Gaussian function, 
\begin{equation}
	w_{ij} = \rm{exp} (-\frac{\emph{k}^{2}}{2\emph{b}^2})
\end{equation}
where \emph{b} is the Gaussian bandwidth.  The attenuated coefficient $w_{ij}$ ensures that entities $e_j$ closer to $e_i$ is assigned a higher weight.

Then, we employ the activation function LeakyRelu to calculate the importance of each triplet $\emph{t}_{ikj}$ in $\mathcal{G}_t$  denoted by $\beta_{ikj}$,   
\begin{equation}
 \beta_{ikj} = \rm{LeakyReLU}(\textbf{W}_2 \textbf{\emph{t}}_{\emph{ikj}})
\end{equation}
where $\textbf{W}_2$ is a linear transformation matrix. 

Afterwards, we further  calculate the attenuated attention weight  $\alpha_{ikj} $  of each triplet in $\mathcal{G}_t$, 
\begin{equation}
	\alpha_{ikj} = \frac{\rm{exp}(\emph{$ w_{ij} \beta_{ikj}$})}{\sum\nolimits_{n \in \mathcal{N}_i}\sum\nolimits_{r \in \mathcal{R}_{in}}\rm{exp}(\emph{$w_{in} \beta_{irn}$})}
\end{equation}
where $\mathcal{N}_i$ represents direct and multi-hop neighbors of entity $e_i$, and $\mathcal{R}$  denotes direct in-flowing relations and multi-hop relation of the entity $e_i$. 

Based on the multi-head attention mechanism that improves the stability of the learning process, we finally update the new embedding of the entity $e_i$  which is the weighted sum of the product between $\alpha_{ikj}$ and $\textbf{\emph{t}}_{ikj}$, 
\begin{equation}
  \textbf{\emph{e}}_i^{s} =  \big \Vert^M_{m=1} \sigma ( \sum\limits_{j \in \mathcal{N}_i }\sum\limits_{k \in \mathcal{R}_{ij}}\alpha_{ikj}^{m} \textbf{\emph{t}}_{ikj}^{m}). 
\end{equation}
Here, $M$ denotes the maximum number of attention heads, $\Vert$ represents concatenation.  $\textbf{\emph{e}}_i^{s}$ is the entity embedding that captures the semantic  dependence of entity $e_i$ in the timestamp $\mathcal{G}_t$. It is fed into the temporal attention network, described below. 

\subsubsection{Temporal Self-Attention Network}
The existing TKG representation methods mostly employ recurrent networks to aggregate history features, but they suffer from information loss and low computing efficiency \cite{attengraph}.  Inspired by the attention mechanism that is efficient and significantly superior in sequential tasks \cite{attention}, we design a temporal self-attention network (TSAN) in DREAM. We define the input of TSAN as a sequence of representations \{$\textbf{\emph{e}}_{i}^{s1}$, $\textbf{\emph{e}}_{i}^{s2}$ ,..., $\textbf{\emph{e}}_{i}^{s(t-1)}$, $\textbf{\emph{e}}_{i}^{st}$\} for a particular entity $e_i$ at timestamp $t$ ($t$<$t_q$), where $e^{s}$ with dimensionality $D^{'}$ has sufficiently captured semantic dependence in RAGA. In addition, the output of TSAN is a new representation sequence \{$\textbf{\emph{z}}^{e_i}_1$, $\textbf{\emph{z}}^{e_i}_2$, ..., $\textbf{\emph{z}}^{e_i}_{t-1}$, $\textbf{\emph{z}}^{e_i}_t$\} for entity $e_i$ at different timestamps, $\textbf{\emph{z}}$ $\in$ $\mathbb{R}^{F^{'}}$ with dimensionality $F'$. Formally, by packing together across timestamps $t$, the representations of input and output are denoted as $\textbf{X}^s$ $\in$ $\mathbb{R}^{t \times D'}$ and $\textbf{Z}_{\emph{t}}^{e_i}$ $\in$ $\mathbb{R}^{t \times F'}$, respectively.


Technically, TSAN's self-attention calculates the similarity between historical events to capture the temporal evolution. Specifically, queries \textbf{\emph{Q}}, keys \textbf{\emph{K}}, and values \textbf{\emph{V}}  are regarded as input entity representations. \textbf{\emph{Q}}, \textbf{\emph{K}}, and \textbf{\emph{V}} $\in$  $\mathbb{R}^{D' \times F'}$ have the same shape by using linear projection matrices $\textbf{W}_q$, $\textbf{W}_k$, $\textbf{W}_v$ to transform queries, keys, and values, respectively. 
  Inspired by previous studies \cite{RE-GCN, RE-NET}, TSAN allows each timestamp $t$ to attend over all timestamps up to  $t$, which preserves the auto-regressive property. 

We first model the correlation of temporal evolution by computing the dot product, and then apply the softmax function to obtain the temporal attention values. Next, the new embedding of the entities before timestamp $t$ is the sum of each representation weighted by their temporal attention values. The above processes are defined as follows,
\begin{equation}
	\textbf{Z} = (\textbf{X}^s \textbf{W}_v) \rm Softmax ((\textbf{X}^{\emph{s}} \textbf{W}_\emph{q})^{\emph{T}}(\textbf{X}^{\emph{s}} \textbf{W}_\emph{k}))
\end{equation}

Finally, we adopt $M'$ attention heads and compose multi-faceted representation of the entity $e_i$ at timestamp $t$, 
\begin{equation}
	\textbf{Z}_{\emph{t}}^{e_i} = [\textbf{Z}_{1}\oplus \textbf{Z}_{2}\oplus... \oplus \textbf{Z}_{M'}]
\end{equation}
where $M'$ is the maximum number of attention heads. The output of TSAN is updated by the multi-head attention  mechanism. Note that, the semantic representation of the entity generated by RAGA is fed into the TSAN module to learn  temporal information. Based on this, MFAR captures both temporal evolution and semantic  dependence simultaneously.

\subsection{Adaptive Reinforcement Learning}

Basically, the existing RL-based TKGR solutions adopt manually designed reward functions,  which leads to  laborious and ineffective reasoning processes \cite{rewardRL, MultiHop, rewardRLcon}. To solve this problem, we propose a novel adaptive reinforcement learning framework (ARLF) in this subsection. The key innovation of ARLF lies in the following two points. (1) For different TKG datasets, ARLF utilizes generative adversarial networks to adaptively learn rewards by imitating demonstration paths, which stabilizes reasoning performance and alleviates  manual intervention. (2) ARLF not only introduces semantic and temporal knowledge into generative adversarial networks, but also explicitly models the training process of the network in TKGR. This provides a new research  perspective for RL-based reasoning methods  for TKG.

ARLF consists of three modules, namely \emph{RL-based reasoner}, \emph{time-constrained demonstration sampler} and \emph{rule-aware discriminator}. Specifically, the reasoner first leverages an agent to output diverse generated paths about missing elements. Next, a time-constrained demonstration sampler simultaneously selects prior and posterior neighbors to generate high-quality demonstration paths. Then, the rule-aware discriminator distinguishes these paths from both semantic and temporal rule levels. Finally, to gain more adaptive reward from the rule-aware discriminator, the reasoner tries to generate paths to deceive the rule-aware discriminator by imitating the demonstrations. Based on this, RL-based reasoner updates the reasoning policy until missing elements are predicted.

\subsubsection{RL-based Reasoner}
RL-based reasoner regards the reasoning process as the MDP where the goal is to choose a sequence of the optimal actions  
 and update its reasoning policy (i.e., the prediction of missing elements). This interaction step  is as intuitive as “take a walk”, which intuitively shapes an explainable results for TKGR \cite{MINERVA}.  Thus, the reasoner trains an agent to interact with TKGs by a \emph{4}-tuple of MDP (State, Action, Transition, Reward). 

\textbf{State.}  The state consists of some TKG elements to describe the local environment. Formally, each state $s_l$ at reasoning step $l$ is denoted $s_l$ = ($e_l$, $t_l$, $e_s$, $t_q$, $r_q$) $\in$ $\mathcal{S}$, where ($e_l$, $t_l$) represents the entity and timestamp at the current reasoning step $l$, and ($e_s$, $t_q$, $r_q$) corresponds to the entity, timestamp and relation from the original query that remain fixed throughout all steps. The state space $\mathcal{S}$ can be viewed as a set of available states. The initial state is ($e_s$, $t_q$, $e_s$, $t_q$, $r_q$) since the RL-based agent starts
from  $e_s$ of the quadruple query. Considering the continuous evolution of entities over time, the multi-faceted attention representation is applied in the state representation. Thus, $\textbf{Z}_{t_{l}}^{e_l}$ denotes the representation of the entity $e_l$ at the timestamp $t_l$. In addition, following the common practice in TKGR \cite{RE-NET, TKG-survey-bai}, we use $\textbf{Z}_{t_q-1}^{e_s}$ 
 to approximate the multi-faceted representation of entity $e_s$ at future timestamp $t_q$ since $t_q$ is in fact unobservable during inference.


\textbf{Action.} Action space $\mathcal{A}$ for the given $s_l$ includes the set of valid actions $A_l$ at the reasoning step $l$.  $A_l$  is formulated as $A_l$  = \{($r'$, $e'$, $t'$) $\mid$ ($e_l$, $r'$, $e'$, $t'$) $\in$ $\mathcal{A}$,  $t'$  $\leq$ $t_l$, $t'$ $\leq$ $t_q$  \}. In addition, we add a self-loop edge to every $\mathcal{A}$. When reasoning is unrolled to  the maximum reasoning step $L$, the self-loop acts similarly to a $STOP$ action, which avoids infinitely unrolling in reasoning processes.

\textbf{Transition.} The new state $s_{l+1}$ is updated by the previous state $s_l$ and the action $A_l$. Formally, $\mathcal{P}_r$: $\mathcal{S}$ $\times$ $\mathcal{A}$ $\rightarrow$ $\mathcal{S}$ is defined as $\mathcal{P}_r$ (\emph{$s_l$}, \emph{$A_l$}) = $s_{l+1}$= ($e_{l+1}$, $t_{l+1}$, $e_q$, $t_q$, $r_q$), where $\mathcal{P}_r$ is the transition function. 

\textbf{Reward.}  We define a coarse-grained terminal reward to guide the agent select a basic policy so that it can be close to
 the target entity $e_d$ during initial training.  This reward is a basic  reward paradigm for all reinforcement learning methods \cite{rewardRL}. The terminal reward $R_t$ is set to 1 if the target entity is the ground truth entity $e_d$. Otherwise, the value of the terminal reward is 0. Note that, the final reward $R(s_l)$ is defined in Eq. (18).

\textbf{Policy Network.} 
The goal of the reasoner is to learn a policy network that drives the interaction between the aforementioned \emph{4}-tuple of MDP and TKGs. This policy network inputs the embedding learned from MFAR, and outputs the next action. Unlike single-hop methods, multi-hop reasoning methods keep a reasoning history. Formally, the reasoning history $h_l$ consists of the sequence of visited  elements, i.e., ($e_s$, $t_q$), $r_1$,  ($e_1$, $t_1$), ..., $r_l$, ($e_l$, $t_l$). We define an improved path representation $\textbf{P}$ to concatenate the relation embedding and the entity embedding $\textbf{P}_l$ = [$\textbf{R}_l$ $\oplus$ $\textbf{Z}_{t_{l}}^{e_l}$] by reshaping  $\textbf{r}_l$ $\in$ $\mathbb{R}^{F'}$ to $\textbf{R}_l$ $\in$ $\mathbb{R}^{1 \times F'}$. The history embedding for $h_l$ can be encoded using Long Short-Term Memory (LSTM),
\begin{equation}
	\textbf{H}_l= \rm{LSTM}([\textbf{H}_\emph{l-1}\oplus \textbf{P}_\emph{l-1}]).
\end{equation}

Based on this,  policy network $\pi$ that calculates the probability distribution of the action space can be defined as follows,
\begin{equation}
	\pi_\theta(\emph{a}_l|\emph{s}_l) = \sigma(\textbf{A}_{l}(\textbf{W}^{''}\text{ReLu}(\textbf{W}^{'}[\textbf{Z}_{t_{l}}^{e_l}\oplus\textbf{H}_{l}\oplus\textbf{R}_{q}])))
\end{equation}
where $\mathcal{A}_l$ is encoded as $\textbf{A}_{l}$ by stacking the embeddings of all actions, $\sigma$ is the softmax operator to maximize the probability of the next action in the action space $\mathcal{A}_l$.

\subsubsection{Time-constrained Demonstration Sampler}

In this subsection, we propose a time-constrained demonstration sampler to extract high-quality demonstrations (expert paths) considering the temporal property of TKGs. Specifically, we use bi-directional breadth-first search (Bi-BFS) to explore the shortest paths between  $e_s$ and  $e_d$ for each fact as a  demonstration. 
This is because the combination of the shortest relation path can accurately represent the semantic link between two entities \cite{DeepPath, MultiHop, kgr_survey}. However, traditional Bi-BFS uniformly samples prior neighbors and posterior neighbors, which may accidentally explore  neighbors beyond the reasoning time range. To address this problem, we design an exponentially weighted mechanism into bi-directional breadth-first search . For example, considering sampling prior neighbors in the Bi-BFS, the sampling probability of exponentially weighted mechanism is defined  as the following formula,
 \begin{equation}
 	\mathcal{P}(e_{i}, r_i, e_j, t') =  \frac{\rm{exp}(\emph{$t' - t$})}{\sum\nolimits_{(e_i, r_l, e_k, t'') \in \mathcal{N}_q}\rm{exp}(\emph{t}''-\emph{t})}
 \end{equation}
where $\mathcal{N}_q$ represents prior neighbors of $e_i$. $\emph{t}''$ and $\emph{t}'$ are prior to $\emph{t}$. Essentially, with exponential weighting, prior entities from a timestamp closer to $t$ will be assigned a higher probability during sampling. 

\subsubsection{Rule-aware Discriminator}


Inspired by \cite{DIVINE}, rule-aware discriminator is designed to evaluate  generated paths obtained by the reasoner and demonstrations from the semantic and temporal logic levels respectively.  The rule-aware discriminator includes a semantic discriminator and a temporal rule discriminator.

\textbf{Semantic Discriminator.} To distinguish the difference between the demonstration path and the generated path at the semantic level, the discriminator first learns the representation \textbf{P} of  path by adopting the path concatenation 
 mentioned in the policy network. Inspired by ConvKB that demonstrated effectiveness of convolutional neural networks in extracting semantic features of KGs \cite{ConvKB}, we employ a convolutional layer by sliding a kernel $\textbf{\emph{w}}$ to extract the semantic feature of embedding \textbf{P},
 \begin{equation}
	D(s) = \sigma (\textbf{W}_s(\rm ReLu (Conv(\textbf{P}, \textbf{\emph{w}}) + \textbf{\emph{b}}_s))
\end{equation}
where $\textbf{\emph{b}}_s$ denotes a bias term, and the sigmoid function $\sigma$ outputs the path semantic feature $D(s)$ on the interval (0,1). 

\textbf{Temporal Logic Discriminator.} 
To improve the transparency and trustworthiness of paths, the temporal rule discriminator further learns the temporal logical rules from TKGs, so that the reasoner can directly imitate the temporal logic implied in the time-constrained demonstration. 
A temporal logical rule $R_t$ is defined as, 
 \begin{equation}
	\wedge_{i=1}^{l}(E_i, r_i, E_{i+1}, T_i) \Rightarrow (E_1, r_h, E_{l+1}, T_{l+1})  
\end{equation}
where $\wedge$ is logic conjunction, $T_1<T_2<...<T_l<T_{l+1}$. The right-side of $R_t$ is called a rule head, $r_h$ denotes the head relation. In addition, the left-side of $R_t$ is  a rule body, which is represented by a conjunction of atomic quadruple ($e_s$, $r_1$, $e_2$, $t_1$) $\wedge$ ... $\wedge$ ($e_i$, $r_i$, $e_{i+1}$, $t_i$) $\wedge$ ... $\wedge$ ($e_l$, $r_{l-1}$, $e_{l-1}$, $t_{l-1}$) $\wedge$ ($e_{l-1}$, $r_{l}$, $e_{l}$, $t_{l}$). We use $\mathcal{Q}_1$, ..., $\mathcal{Q}_l$ to denote the above quadruples, and $\mathcal{Q}_h$ denotes a rule head.

To add the rationality of temporal rules hidden in paths, we employ the \emph{t}-norm fuzzy logic \cite{fuzzylogic}, which introduces a truth measure of rules as the combination of the truth measures in the rule body.  To obtain the truth measure  $I$($\mathcal{Q}_i$) of a  quadruple, we first obtain the concatenation embedding of the relation and entities, and then employ a multi-layer neural network as the feature extractor, which is represented as follows, 
 \begin{equation}
I(\mathcal{Q}_i) = \sigma(\textbf{W}_{r}tanh([\textbf{Z}_{t_{i}}^{e_i} \oplus\textbf{\emph{R}}_{i}\oplus \textbf{Z}_{t_{i}}^{e_{i+1}}))
\end{equation}

Based on this, we leverage \emph{t}-norm fuzzy logic to formulate the feature of temporal logic for  paths as follows:
 \begin{equation}
	D(r) = I(\mathcal{Q}_1)\cdot...\cdot I(\mathcal{Q}_l) \cdot I(\mathcal{Q}_h) - I(\mathcal{Q}_1) \cdot...\cdot I(\mathcal{Q}_l)+1 
\end{equation}

\subsubsection{Training}
In this section, we follow the existing work \cite{RIDIVINE} and define the final reward function and formalize the training process. Specifically, we first introduce path noise $D^N(s)$ and logic noise $D^N(r)$ with uniform distribution to discard low-quality paths obtained from RL-based reasoner, and obtain the adaptive reward from semantic and temporal logic levels. 
 \begin{equation}
	R_s = \max(D^G(s) - D^N(s), 0)
\end{equation}
 \begin{equation}
	R_r = \max(D^G(r) - D^N(r), 0)
\end{equation}
where $D^G(s)$ and $D^G(r)$ are semantic feature and temporal logic feature of reasoning path obtained by RL-based reasoner. $R_s$ and $R_r$ represent semantic reward and temporal logic reward, respectively.

Next, we define the adaptive reward $R(s_l)$ at the state $s_l$ as follows:
 \begin{equation}
	R(s_l) = R_t+\alpha R_s + (1-\alpha) R_r
\end{equation}
where $\alpha$ is a balance factor to weight $R_s$ and $R_r$. Note that, the agent can obtain the adaptive reward at every reasoning step by imitating demonstrations from both semantic and rule levels, which eliminates the sparse reward and ensures performance stability \cite{IRL}. In addition, ARLF improves generalizability and avoids decision bias in different datasets by automatically mining common meta-knowledge from diverse  paths \cite{RLbalan}.

Then, we optimize the semantic and temporal logic discriminators by reducing training loss, which enhances the effectiveness of imitation learning. 
 \begin{equation}
	\mathcal{L}_p = D^G(s) - D^D(s)+ \lambda (\parallel \bigtriangledown_{\hat{\emph{p}}}D(\hat{\emph{p}})\parallel_{2} -1)^{2}
\end{equation}
where $\lambda$ is a penalty term and $\hat{\emph{p}}$ is sampled uniformly along straight lines between the generated path and the expert demonstration \cite{RIDIVINE}. $D^G(s)$ and $D^D(s)$ are semantic features  extracted from generated paths and expert demonstrations, respectively.

For the temporal logic discriminator, we define loss $\mathcal{L}_r$  as follows:
 \begin{equation}
	\mathcal{L}_r = -(\log D^D(r) +\log(1-D^G(r))
\end{equation}
where $D^G(r)$ and $D^D(r)$ are temporal logic features extracted from generated path and the expert demonstration, respectively.

Finally, to maximize the accumulated rewards of adaptive reinforcement learning and obtain the optimal policy, the objective function is as follows,
\begin{equation}
\emph{J}(\theta) = \mathbb{E}_{(e_s,r_q,e_d,t_q)} \mathbb{E}_{a_1,...,a_{L}\sim{\emph{$\pi$}_\theta}}[\emph{R} (s_l\mid\emph{e}_s,r_q,t_q)]
\end{equation}

\section{Experiments}

\subsection{Experimental Setup}
\subsubsection{Datasets.}
We report the reasoning performance of DREAM on four public datasets: ICEWS14, ICEWS18, ICEWS05-15 and GDELT. The first three are subsets of events in the Integrated Crisis Early Warning System (ICEWS) that occurred in \emph{2014}, \emph{2018} and \emph{2005-2015}, respectively. The last one is extracted from the Global Database of Events, Language \cite{gdelt}. We follow \cite{TiTer} to split datasets  and split each dataset into train/valid/test by timestamps.  Numerical information of datasets is shown in Table 1, $Time$ represents time granularity. 

\subsubsection{Evaluation metrics.}
We evaluate reasoning performance by adopting two widely used metrics: MRR and  
 Hits@N. Note that, to report our experimental results, we employ the more reasonable \emph{time-aware filtered setting} \cite{TiRGN} rather than the  \emph{filtered setting} \cite{HIP, CENET} or \emph{raw setting} \cite{rGalT, DA-NET}. This is because the \emph{time-aware filtered setting} only filters out the quadruples occurring at the query time \cite{HiSMatch}. In addition, \emph{time-aware filtered setting} can simulate extrapolation prediction tasks in the real world  \cite{TiTer}.

\begin{table}
	\centering
 \small
	\caption{Statistics of the Experimental Datasets.}
 \vspace{-0.2cm}
	\begin{tabular}{lllllll}
		\hline
		Dataset     & \#Train & \#Valid & \#Test &\#Ent &\#Rel  &\emph{Time}\\
		\hline
		ICEWS14       & 74,845  & 8,514     &7,371 &6,869 & 230 & 1 day\\
		ICEWS18       & 373,018 & 45,995      & 49,545 &23,033 & 256&  1 day\\
         ICEWS05-15  & 368,868 & 46,302     & 46,159 &10,094& 251&  1 day\\
         GDELT       & 1,734,399 & 238,765     & 305,241&7,691& 240& 15min\\

		\hline
	\end{tabular}
	\vspace{-0.4cm}
	\label{tab:plain}
\end{table}

\begin{table*}
  \label{table1}
	\centering
	\caption{Results of Link Prediction on Public TKGs.}
 \vspace{-0.35cm}
	\setlength{\tabcolsep}{0.7mm}{
		\begin{tabular}{c|cccc|cccc|cccc|cccc}
			\toprule
			&  \multicolumn{4}{c|}{ICEWS14} & \multicolumn{4}{c|}{ICEWS05-15} &  \multicolumn{4}{c|}{ICEWS18} & \multicolumn{4}{c}{GDELT}\\ 
			Model         & MRR    & H@1 & H@3    &H@10    & MRR        & H@1   & H@3    & H@10  & MRR        & H@1   & H@3    & H@10 & MRR        & H@1   & H@3    & H@10  \\ \midrule
 TA-DistMult   & 25.8 & 16.9 &29.7 &43.0& 24.3& 14.6 & 27.9& 44.2 & 16.7 & 8.6&18.4&33.6& 12.0& 5.8& 12.9 & 23.5\\	
 TNTComplEx  & 34.1 & 25.1&38.5 &50.9& 27.5&9.5& 30.8 & 42.9 & 21.2 & 13.3&24.0 &36.9& 19.5& 12.4& 20.8 & 33.4\\	
   \midrule
  CyGNet   & 35.1 & 25.7&39.0 &53.6& 36.8& 26.6 & 41.6  & 56.2 & 24.9  & 15.9&28.3 &42.6& 18.5& 11.5& 19.6 & 32.0\\	
   
   RE-NET     & 36.9     & 26.8  &39.5 & 54.8 &43.3& 33.4 & 47.8 & 63.1& 28.8   & 19.1 & 32.4 & 47.5 & 19.6 &12.4 & 21.0 & 34.0\\

    RE-GCN     & 40.4 & 30.7& 45.0 & 59.2& 48.0 &37.3 & 53.9  & 68.3 & 32.6& 22.4 &36.8  & 52.7 & 19.8 & 12.5 &21.0  & 34.0\\
   
			TANGO     & 36.8   &27.3 & 40.8 & 55.1 & 42.9 & 32.7& 48.1& 62.3 & 28.9&19.4 & 32.2&47.0& 19.2 & 12.2& 20.4 & 32.8\\

			XERTE     & 40.0  & 32.1 & 44.6 & 56.2 & 46.6  & 37.8 & 52.3  & 63.9& 30.0  & 22.1& 33.5 & 44.8& 18.9 & 12.3 & 20.1  &30.3 \\
   
			TLogic     & 42.5 & 33.2 & 47.6 &60.3& 47.0& 36.2 & 53.1 & 67.4& 29.6& 20.4 &33.6 & 48.1& 19.8 & 12.2& 21.7& 35.6 \\

         
            Hismatch    & 46.4  & {\underline {35.9}} & {\underline {51.6}}  & 66.8& {\underline {52.9}}  & {\underline {42.0}} & {\underline {59.1}} & {\underline {73.3}}& 34.0 & {\underline {23.9}} & {\underline {37.9}} & 53.9 & {\underline {22.0}}& {\underline {14.5}}& {\underline {23.8}} & {\underline {36.6}}\\

        \midrule
          TPath   & 38.9     &29.4 & 42.5  & 54.6 &45.7 & 36.8 &51.6 & 63.8& 27.4  & 21.1& 31.3 &42.8 & 16.8  & 11.0 & 18.8 & 29.4 \\
          TAgent   &39.1     & 30.7 & 43.2 & 55.2 & 45.1 & 36.3 & 51.0  & 63.4 & 27.2    & 20.1 & 30.7  & 42.3 & 17.7  & 11.8 & 19.6  &30.6\\
          TITer  & 40.9   & 32.1 & 45.5  & 57.6 & 47.7 & 38.0 & 52.9& 65.8& 30.0& 22.1& 33.5 & 44.8& 20.2& 14.1& 22.2 & 31.2\\
         CluSTer  & {\underline {47.1}}   & 35.0 & --  & {\underline {72.0}} & 45.4 & 34.3 & --& 67.7 & {\underline {34.5}}& 22.9& -- & {\underline {57.7}}& 18.5& 12.1& -- & 32.1\\
		\bottomrule
	DREAM & \textbf{51.7} & \textbf{42.0} & \textbf{56.4}  & \textbf{72.4} & \textbf{56.8}  & \textbf{47.3} & \textbf{65.1}  & \textbf{78.6} & \textbf{39.1}   & \textbf{28.0} & \textbf{45.2}  & \textbf{62.7} & \textbf{28.1}  & \textbf{19.3} & \textbf{31.1} & \textbf{44.7} \\
			\bottomrule
		\end{tabular}
	}
	\label{tab:whoiswho}
	\vspace{-3pt}
\end{table*}

\subsubsection{Baselines.}
Existing extrapolated TKGR studies \cite{XERTE, RE-GCN, TiTer} have fully proved that the performance of  traditional KG reasoning methods  (e.g., TransE \cite{TransE}) is lower than that of TKGR methods. Thus,  both interpolated and extrapolated TKGR models are considered in our experiments. Interpolated TKGR models include  TA-DistMult \cite{TA-transE} and TNTComplEx \cite{TNTComplEx}. In addition, we compare our model with the latest models of extrapolated TKGR studies, including non-RL-based models TANGO \cite{TANGO}, XERTE \cite{XERTE}, CyGNet \cite{CyGNet}, RE-GCN \cite{RE-GCN}, RE-NET \cite{RE-NET}, TLogic\cite{TLogic},  and Hismatch \cite{HiSMatch}. Finally, the RL-based extrapolated TKGR methods are set as the third group of baselines that include Tpath \cite{TPath}, TAgent \cite{TAgent}, TITer \cite{TiTer}, and CluSTer\cite{Cluster}.

\subsubsection{Implementation Details.}
There are some hyperparameters in our proposed model. The dimensions $D$ and $F'$ are set to 200. $\lambda$ is set to 5 in all datasets. The number of attention heads $M$ and $M'$ are both 8 on different datasets. For the model training, the maximum training epoch and the size of the convolution kernel are 400 and 3$\times$5, respectively. Detailed discussions on the effects of the  reward balance factor $\alpha$, the maximum reasoning step $L$, and the Gaussian bandwidth $b$ will be shown in the subsection \textbf{Sensitivity Analysis}.


\subsection{Results on TKG Reasoning}
Reasoning performance are reported in Table 2  (these scores are in percentage), where the competitive results of baseline are marked by underlining and the highest performance results are highlighted in bold. We observe that DREAM consistently surpasses all the baselines on all metrics and datasets, which justifies its  superiority in reasoning performance. Specifically, we have the following detailed observations. (1) Extrapolated TKGR models generally perform better than the first group of interpolated TKGR baselines. We believe that this result occurs because interpolated TKGR models ignore the importance of temporal history  for predicting future events \cite{TiTer}. 
(2) Our proposed model outperforms the second group of extrapolated baselines in all cases. This is because DREAM obtains multi-faceted representation by capturing both temporal evolution and semantic dependence jointly. Moreover, RL-based DREAM can leverage the optimal reasoning policy to arrive correct target by multi-step inferring  among different reasoning clues. (3) Compared with the state-of-art RL-based TKGR methods, the performance improvement of our model on the ICEWS05-15 is more obvious than that on other datasets. 
(4) Since entities are abstract concepts, the reasoning performance of all models on GDELT is generally low \cite{Cluster}, but our model still maintains competitive performance. This is because DREAM captures multi-faceted representations to adapt to frequently changing historical facts on GDELT that has the most fine-grained time granularity \cite{HiSMatch}.

\subsection{Ablation Study}

To study how each component of DREAM contributes to the performance, we perform an ablation analysis. Table 3   reports the MRR results of the DREAM and its variants. The first variant version w/o MFAR only leverages static representations. w/o  RAGA only uses the temporal attention while w/o TSAN only considers semantic dependence in the multi-faceted representation method. Note that, to evaluate the effectiveness of ARLF,  \emph{w/o ARLF} removes ARLF and retains only the RL-based reasoner with basic terminal rewards.  

\begin{table}
\centering
\small
	\caption{Ablation Study on Four Datasets. }
 \vspace{-0.35cm}
	\begin{tabular}{lllll}
		 \hline
		Model     & ICEWS14 & ICEWS05-15 & ICEWS18 &GDELT \\
		\hline
	w/o MFAR & 39.8(-11.9) & 46.2(-10.6) & 28.0(-11.1) &17.5(-10.6)\\
        w/o  RAGA & 49.5(-2.2) & 53.4(-3.4) & 36.8(-2.3) &26.0(-2.1)\\
        w/o TSAN & 46.2(-5.5) & 51.0(-5.8) & 33.5(-5.6) &21.8(-6.3)\\
        w/o ARLF & 47.4(-4.3) & 52.3(-4.5) & 34.9(-4.2) &23.7(-4.4)\\
         \hline
         DREAM   & \textbf{51.7} & \textbf{56.8} &\textbf{39.1} &\textbf{28.1}\\
	 \hline
	\end{tabular}
	\label{tab:plain}
 \vspace{-0.3cm}
\end{table}

\begin{figure*}
\begin{minipage}[t]{0.22\textwidth}
\centering
\includegraphics[width=\textwidth,height=2.7cm]{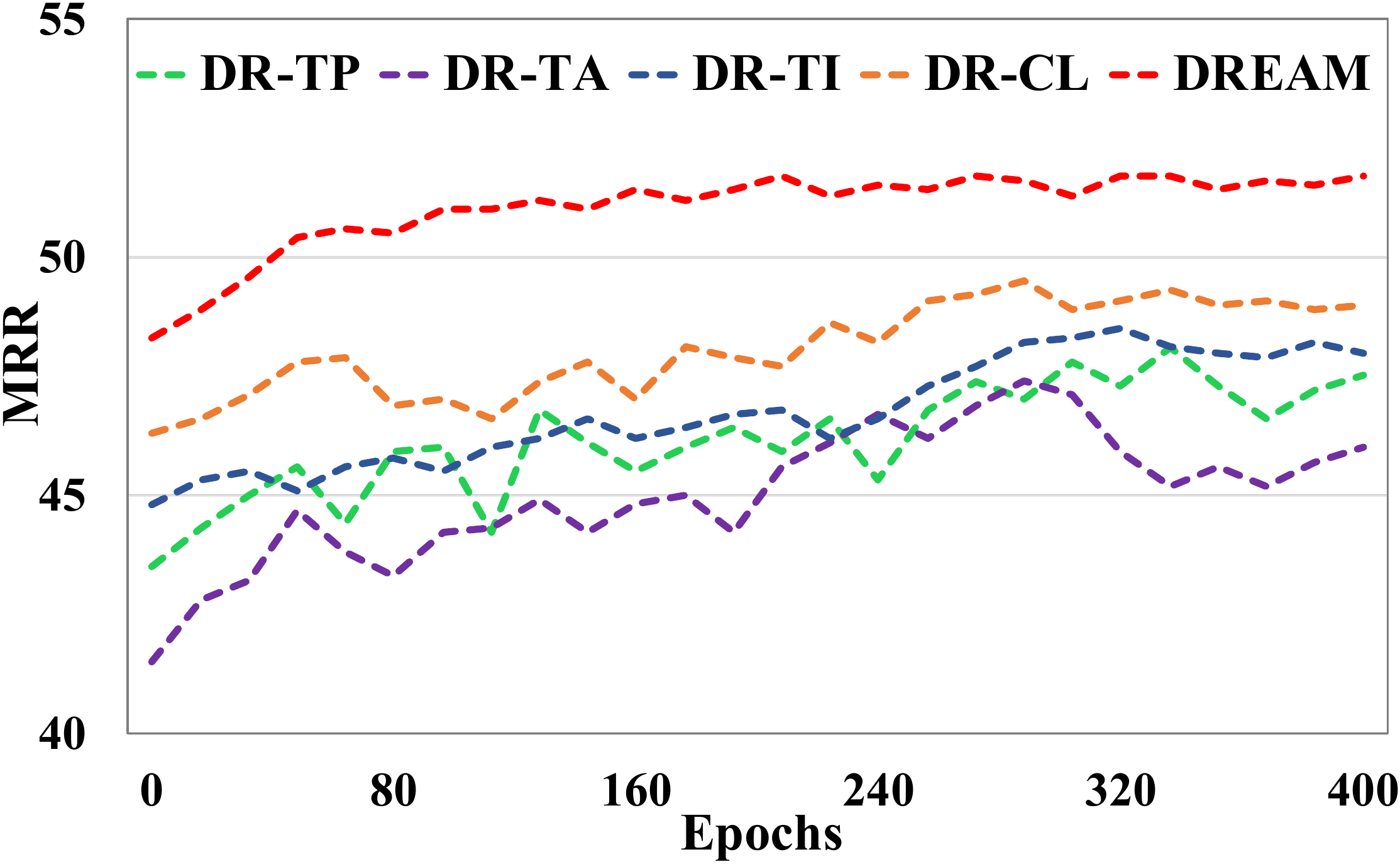}
\subcaption{ICEWS14}

\end{minipage}
\begin{minipage}[t]{0.22\textwidth}
\centering
\includegraphics[width=\textwidth,height=2.7cm]{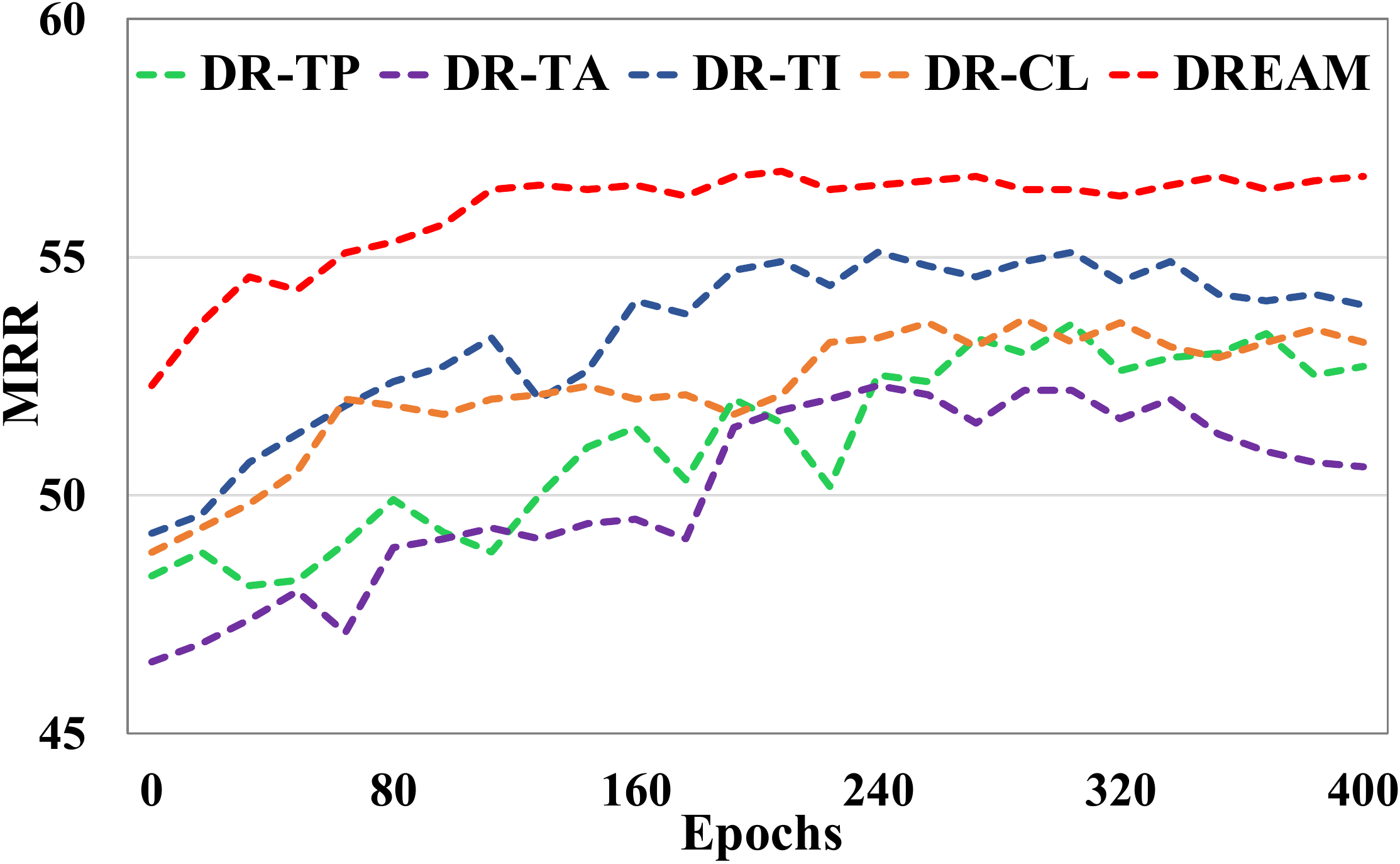}
\subcaption{ICEWS05-15}

\end{minipage}
\begin{minipage}[t]{0.22\textwidth}
\centering
\includegraphics[width=\textwidth,height=2.7cm]{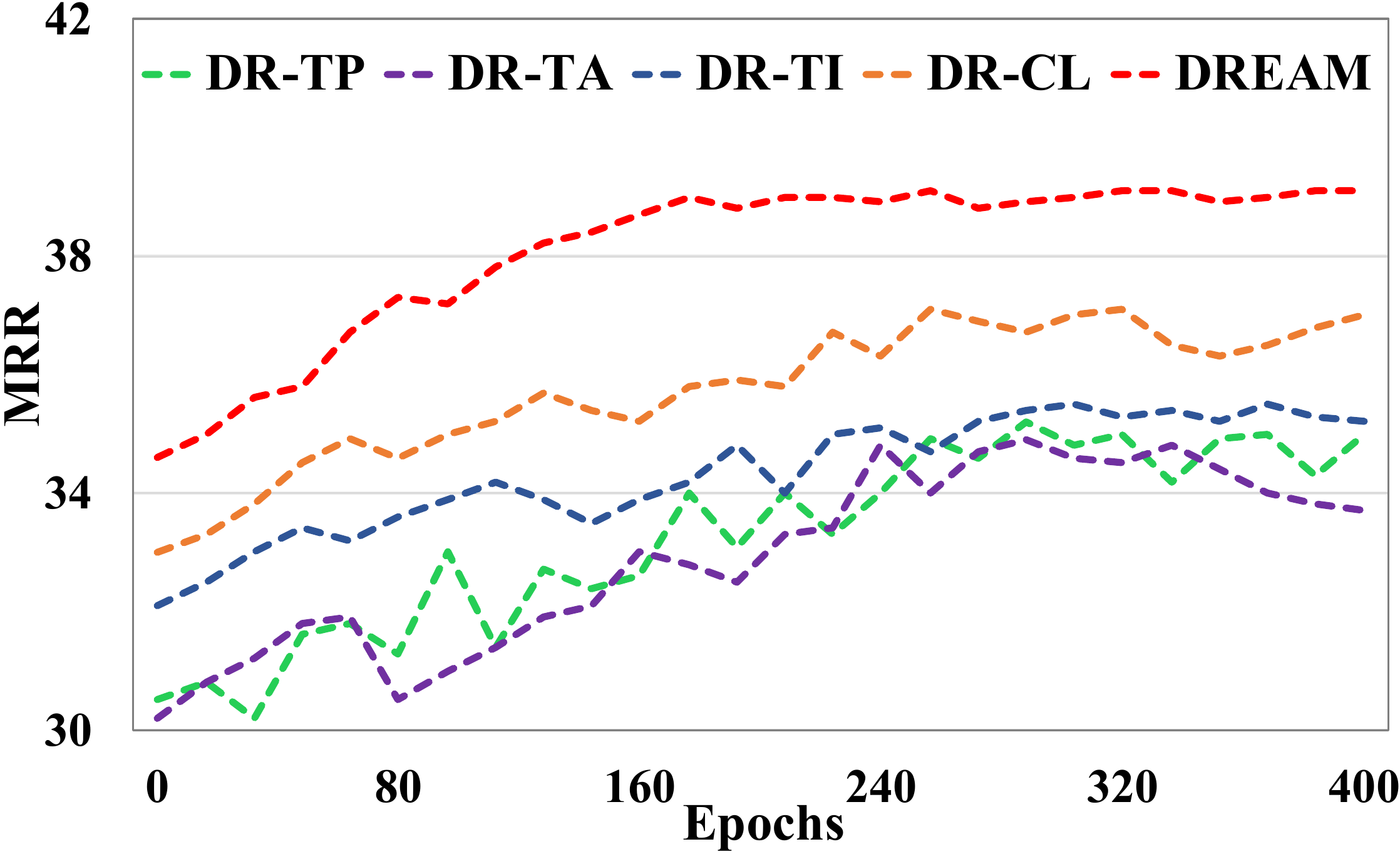}
\subcaption{ICEWS18}

\end{minipage}
\begin{minipage}[t]{0.22\textwidth}
\centering
\includegraphics[width=\textwidth,height=2.7cm]{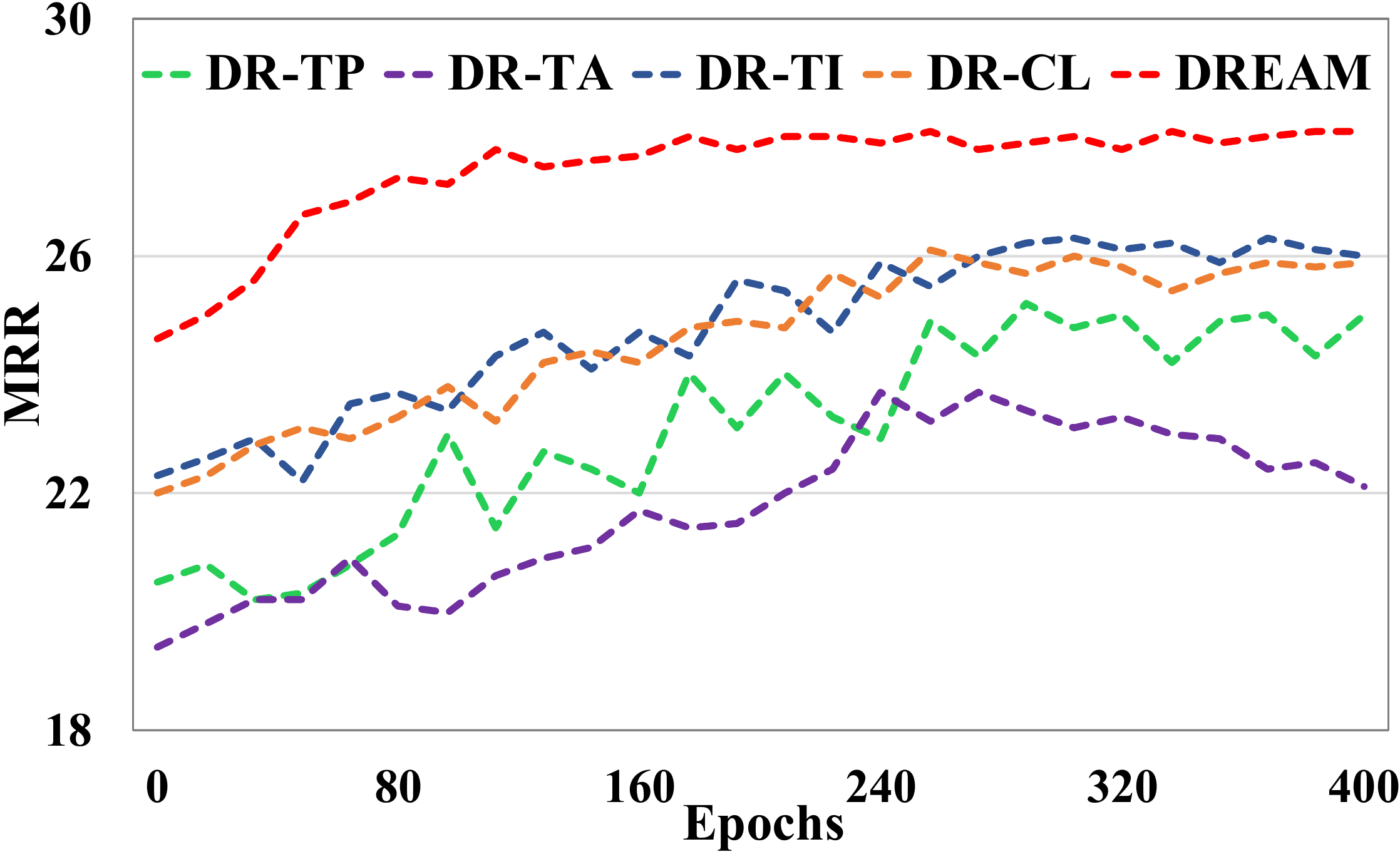}
\subcaption{GDELT}
\vspace{-0.15cm}
\end{minipage}

\caption{ The performance of DREAM and its variant models with different reward functions on datasets.}\label{FIGURE3}
\end{figure*}

We have the following observations. (1) The MRR value of \emph{w/o MFAR} is much lower than that of DREAM, which verifies the necessity of capturing temporal evolution and semantic dependence simultaneously. (2) \emph{w/o  RAGA} cannot leverage semantic dependence, which causes a relatively significant drop in reasoning performance on the sparse dataset, i.e. ICEWS05-15. (3) \emph{w/o TSAN} has a performance drop of more than 5\% on all datasets, which demonstrates the effectiveness of TSAN in improving reasoning performance on TKGs. There are two reasons for the above effectiveness. The first is that the objective existence of temporal evolution between events is the theoretical basis for improving reasoning performance in TKG. Second, query-related entities whose time intervals do not exceed 60 account for 92\% and 83\% on ICEWS subsets and GDELT, respectively. This ratio of the sequence lengths ensures the technical effectiveness of self-attention mechanisms on existing benchmark datasets. 
(4) Although the variant \emph{w/o ARLF} adopts multi-faceted representations to conduct TKGR, its performance still drops on all datasets. This experimental result can explain why existing RL-based TKGR models pursue a reasonable reward mechanism. A detailed investigation of reward mechanisms is presented below.

\subsection{Convergence Analysis}
To analyze the convergence rate and reasoning performance between our adaptive reward and manual rewards in existing RL-based TKGR models, we design four variants (i.e., DR-TP, DR-TA, DR-TI and DR-CL) by removing adaptive rewards and adding manual rewards in DREAM. Similar to  TAgent \cite{TAgent} and TPath \cite{TPath}, the reward of DR-TA only includes binary terminal rewards while DR-TP additionally adds path diversity reward. In addition, DR-CL and DR-TI add the beam-level reward of CluSTer \cite{Cluster} and the time-shaped reward of TITer\cite{TiTer}, respectively. The convergence rate is shown in Figure 3. DREAM adopted the adaptive reward has the fastest convergence and the most stable performance on different TKG datasets. This is because the adaptive reward function learned from both semantic and temporal levels can eliminate the dilemma of sparse rewards and avoids decision bias on different TKG datasets \cite{RLregener}. DR-TA suffers from sparse reward so the convergence results are the worst. Furthermore, the fluctuation of DR-TP is obvious on different datasets due to the blind exploration caused by manually designed diversity rewards. Although DR-CL and DR-TI can finish convergence slowly, their performance is unstable and low generalizability on all  datasets. In brief, the adaptive reward mechanism of DREAM not only converges efficiently, but also has good generalizability on all datasets.

\subsection{Sensitivity Analysis}
In this subsection, we investigate the performance fluctuations of DREAM with varied hyperparameters. Particularly, as mentioned in the subsection \emph{Implementation Details}, we study our model’s sensitivity to the value of Gaussian bandwidth $b$,  the maximum  reasoning step $L$,  as well as the reward balance factor $\alpha$ in Figure 4. In addition, according to the definition of Eq. (3), the value of bandwidth and attenuated weight are negatively correlated. We observe that the optimal bandwidth value on GDELT is smaller than that on all ICEWS subsets. One potential reason is that dense data in GDELT  requires a large attenuation weight to filter noise from neighbors. If the value of $b$ exceeds the optimal value, the values  of MRR will be relatively stable on all datasets. This is because the range of the local influence of the Gaussian kernel and the bandwidth value are positively correlated. Beyond this range,   the value of the kernel function changes slightly, which results in the impact of $b$ on the attenuation weight becomes stable. As shown in Figure 4(b), the performance of DREAM  declines relatively if the number exceeds 3 hops, which is consistent with the results of existing RL-based TKGR studies \cite{TiTer, Cluster}. A reasonable explanation is that more noise contained in the information beyond 3 hops negatively affects reasoning performance \cite{TiTer}. Figure 4(c) reports that the optimal values of the balance factors are diverse on the four datasets. The optimal balance factors are 0.5, 0.6, 0.6, and 0.4 on the datasets ICEWS14, ICEWS05-15, ICEWS18, and GDELT, respectively. The subsets of ICEWS focus on rewards learned at the semantic level, which indicates the scale of the dataset is positively correlated with the adaptive reward at the semantic level when the time granularity is the same  in TKG datasets \cite{Cluster}.  

\begin{figure}
\begin{minipage}[t]{0.147\textwidth}
\centering
\includegraphics[width=\textwidth,height=2.1cm]{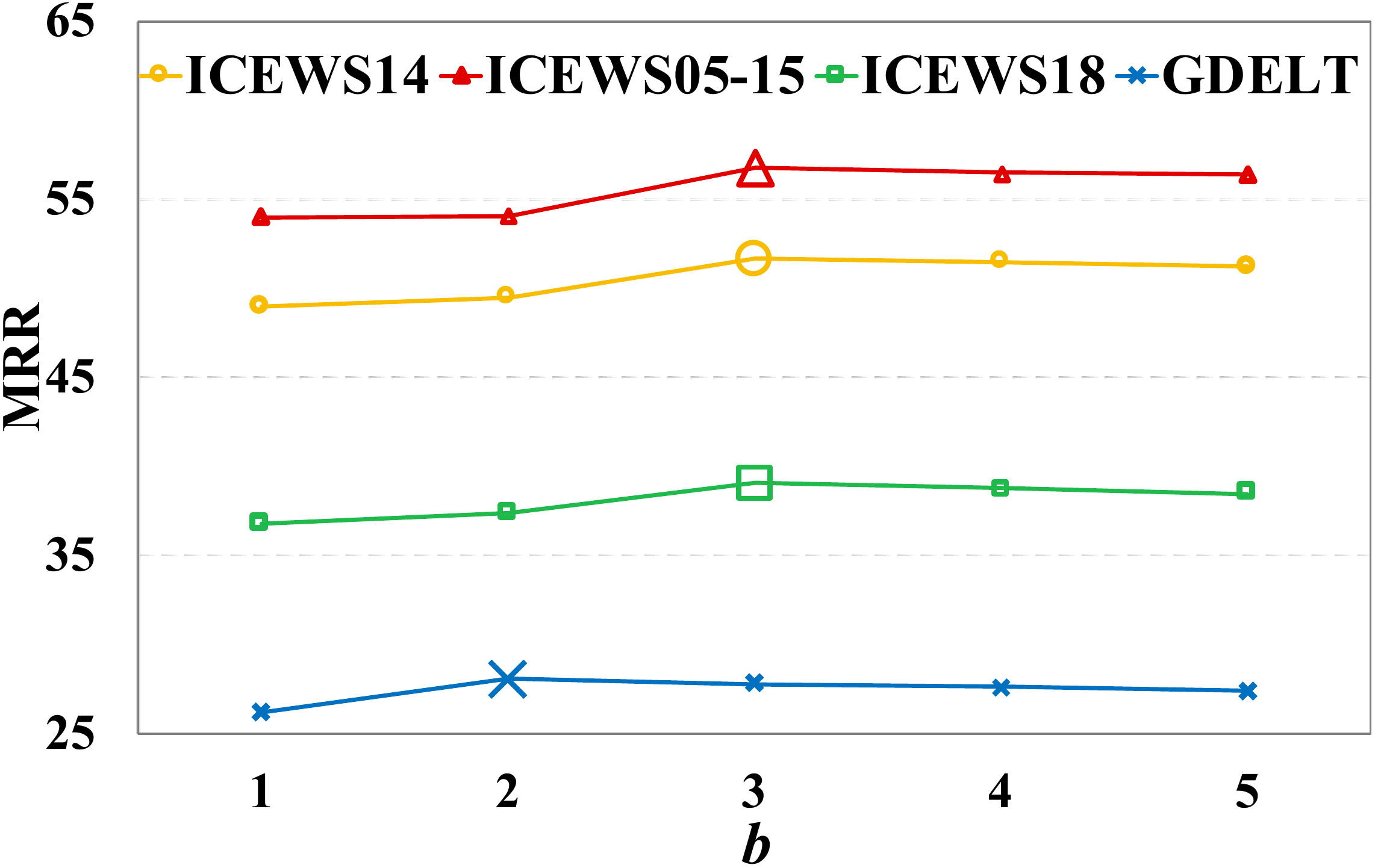}
\subcaption{Bandwidth}

\end{minipage}
\begin{minipage}[t]{0.147\textwidth}
\centering
\includegraphics[width=\textwidth,height=2.1cm]{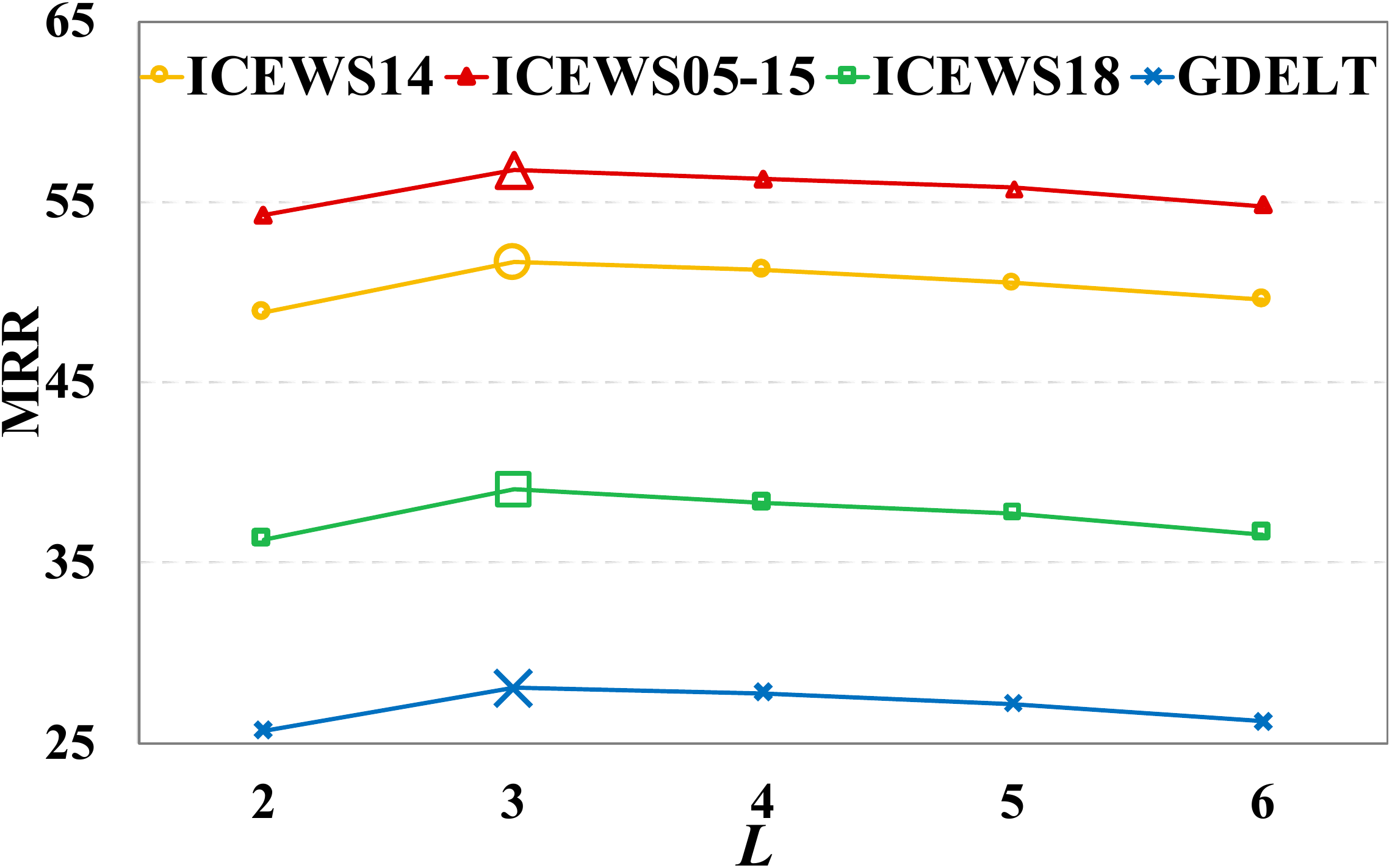}
\subcaption{Reasoning Step}

\end{minipage}
\begin{minipage}[t]{0.147\textwidth}
\centering
\includegraphics[width=\textwidth,height=2.1cm]{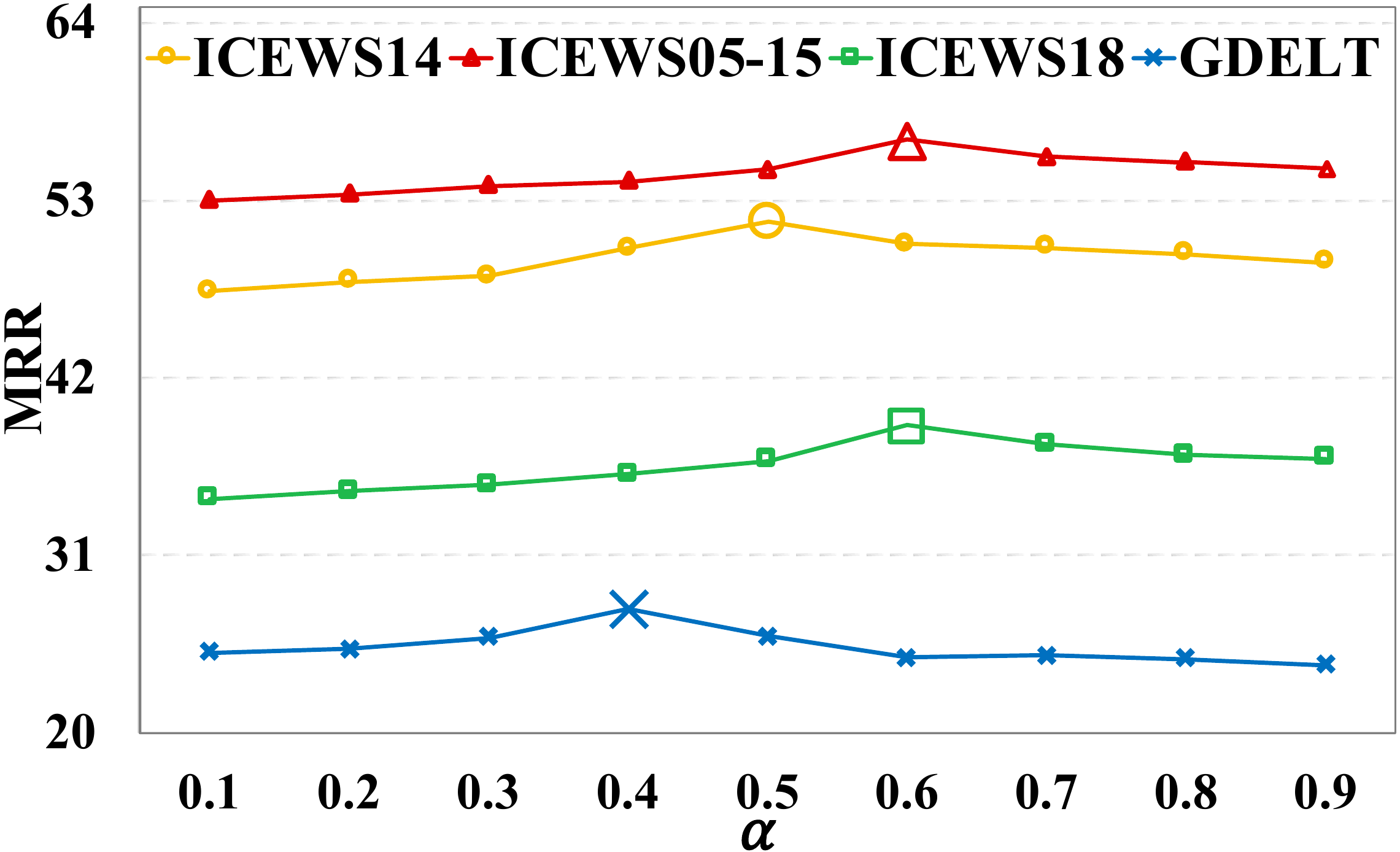}
\subcaption{Balance Factor}

\end{minipage}
\vspace{-0.2cm}
\caption{ The performance of DREAM with the varying number of key hyperparameters on different datasets.}
\label{FIGURE4}
\vspace{-0.56cm}
\end{figure}

\section{Conclusion}
In this paper, we propose a novel RL-based TKGR model entitled DREAM, which overcomes two challenges: the lack of multi-faceted representation and excessive reliance on manually designed reward functions.  For the former, DREAM adopts the MFAR to learn both temporal evolution and semantic dependence. To alleviate reliance on manual rewards, we design a novel ARLF, where the framework is expected to conduct multi-hop reasoning by imitating  expert demonstrations. A large number of experiments demonstrate the reasoning advantages and effectiveness of DREAM on public temporal
knowledge graph datasets. 

\begin{acks}

This work is partially supported by Australian Research Council under the streams of Future Fellowship (No. FT210100624), Discovery Project (No. DP190101985), Discovery Early Career Research Award (No. DE200101465 and No. DE230101033), Industrial Transformation Training Centre (No. IC200100022).  It is also partially supported by National Natural Science Foundation of China (No. 62272332), the Major Program of the Natural Science Foundation of Jiangsu Higher Education Institutions of China (No. 22KJA520006), and National construction of high-level university public graduate project of the China Scholarship Council (No. 202206920032).

\end{acks}

\bibliographystyle{ACM-Reference-Format}
\bibliography{acmart-master/sample-sigconf}


\appendix

\end{document}